\documentclass[12pt,oneside]{CUNY_CS_PhD}
\title{Reservoir Computing: A New Paradigm for Neural Networks}
\author{Felix Grezes}

\usepackage{fullpage}
\usepackage{graphicx}
\usepackage{amssymb}
\usepackage{url}
\usepackage{epstopdf}
\usepackage{enumitem}

\begin{document}

\frontmatter

\maketitle 
\makeabstractpage{Pr. Andrew Rosenberg}{  \textbf{Introduction:}
Even before Artificial Intelligence was its own field of computational science, humanity has tried to mimic the activity of the human brain. In the early 1940s the first artificial neuron models were created as purely mathematical concepts. Over the years, ideas from neuroscience and computer science were used to develop the modern Neural Network. The interest in these models rose quickly but fell when they failed to be successfully applied to practical applications, and rose again in the late 2000s with the drastic increase in computing power, notably in the field of natural language processing, for example with the state-of-the-art speech recognizer making heavy use of deep neural networks.

Recurrent Neural Networks (RNNs), a class of neural networks with cycles in the network, exacerbates the difficulties of traditional neural nets. Slow convergence limiting the use to small networks, and difficulty to train through gradient-descent methods because of the recurrent dynamics have hindered research on RNNs, yet their biological plausibility and their capability to model dynamical systems over simple functions makes then interesting for computational researchers. 

Reservoir Computing emerges as a solution to these problems that RNNs traditionally face. Promising to be both theoretically sound and computationally fast, Reservoir Computing has already been applied successfully to numerous fields: natural language processing, computational biology and neuroscience, robotics, even physics. This survey will explore the history and appeal of both traditional feed-forward and recurrent neural networks, before describing the theory and models of this new reservoir computing paradigm. Finally recent papers using reservoir computing in a variety of scientific fields will be reviewed.
}

\tableofcontents

\mainmatter

\chapter{History of Neural Networks}
\section{Appeal and Historical Recap}
Explaining and reproducing human thought have always interested scientists and philosophers, but only with the discovery of the neuron in the 1890s by Golgi and Ram\'{o}n y Cajal, and with the advent of the computer after 1950 has it finally become a feasible goal. Today artificial intelligence is a major area of research (again to both scientists and philosophers), and artificial neural networks an important paradigm of AI.\\
An early and promising model was the perceptron, proposed in 1957 by Rosenblatt \cite{rosenblatt_perceptron_1957} of the Cornell Aeronautical Laboratory. Modeling a single neuron and using simple algorithms, the perceptron is able to learn a number of functions of its inputs. The output of the perceptron is computed as follows: $f(x) = 1$ if $w \cdot x + b>0$, and $0$ otherwise, where $w \cdot x$ is the dot-product of the input $x$ with the weight vector $w$ (i.e. the weighted sum of the inputs), and $b$ is a bias term which acts a threshold.\\
However in 1969 Minsky and Papert \cite{minsky_perceptrons_1969} showed that not all functions can be learned by the perceptron, most famously the XOR (eXclusive OR) logical operation, since the perceptron is only a linear classifier. Still, research in the field stagnated until Werbos proposed the backpropagation algorithm in 1975, which solved the XOR problem by training over multiple layers of neurons. By the mid 1980s the study of artificial neural networks became a fully established field, with dedicated journals and conferences.

\section{Feed-Forward Networks: Definitions and Theory}
\begin{figure}[!htbp]
\centering
\includegraphics[width=0.9\textwidth]{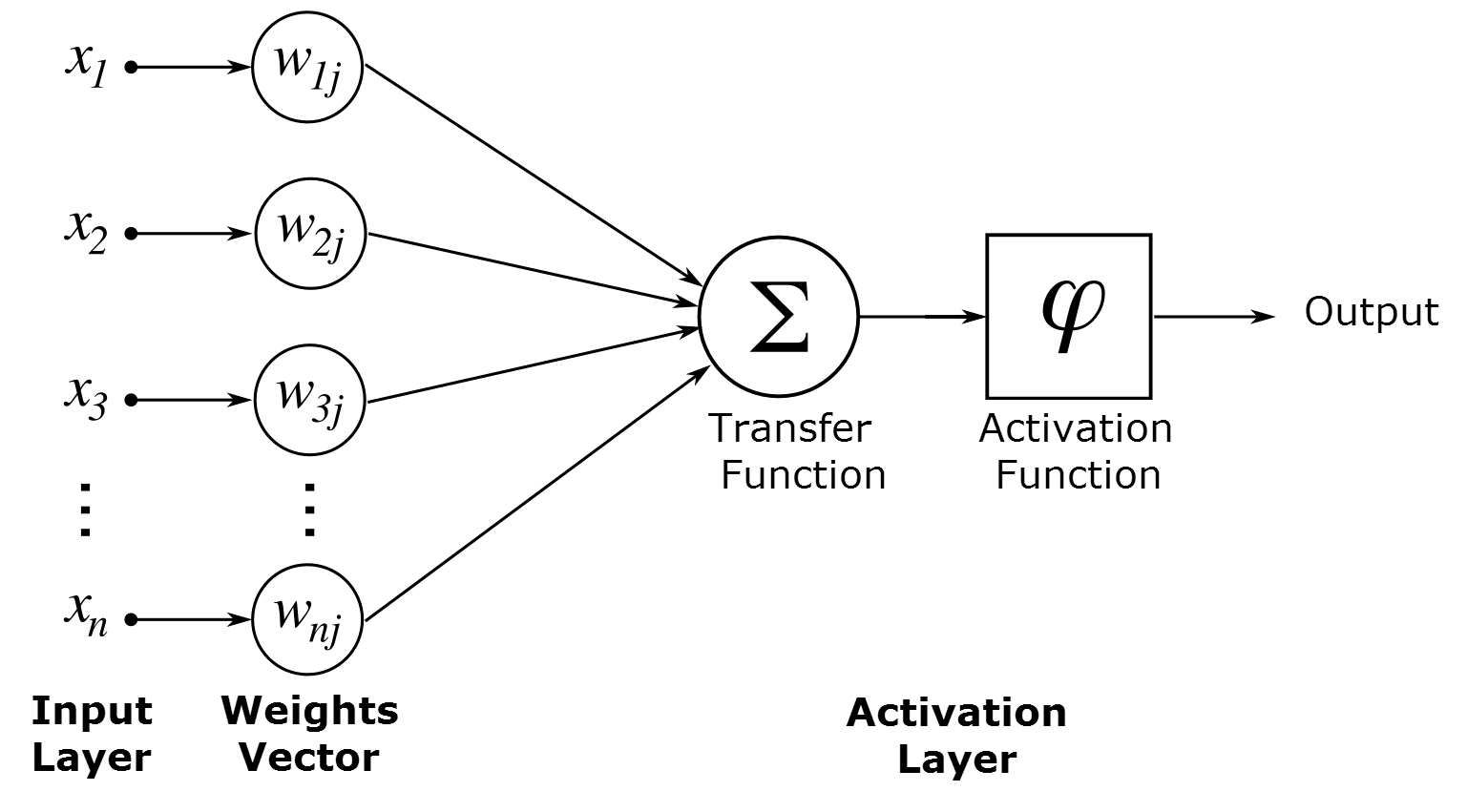}
\caption{Model of an Artificial Neuron}
\label{fig:neuron}
\end{figure}

The fundamental building block of a neural network is the neuron. In essence the neuron is simply a model for a multivariate function whose input variables are weighted by a weight vector. Mathematically: $f(x) = \varphi(w \cdot x +b)$ , with $\varphi$ the chosen activation function, and $w,x,b$ the same as in the perceptron, i.e. respectively the weight vector, the input vector and the bias term. Figure \ref{fig:neuron} gives a visualization of the artificial neuron model.\\

\begin{figure}[!htbp]
\centering
\includegraphics[keepaspectratio=false, height=11.09cm, width=12.5cm]{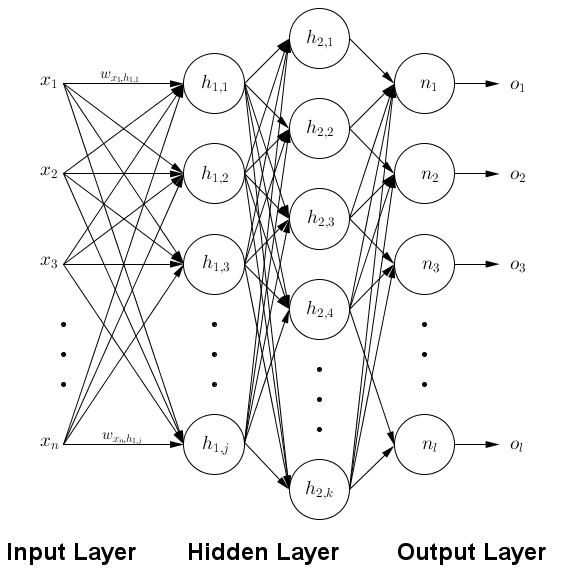}
\caption{Model of an Feedforward Neural Network}
\label{fig:feedforward}
\end{figure}

The activation function is usually chosen to be non-linear, as to allow the neuron, and the networks built from neurons, the power to approximate any mathematical function (see Cybenko-Hornik results below).
Some popular choices are, threshold similarly to the perceptron, tanh and sigmoid which are continuously differentiable. Research is also done on more esoteric activations, such as the cosine or radial functions \cite{lee_cosine-modulated_1996}.\\
However, the perceptron alone can only learn linear functions.  Combining mathematical into networks, similar to how biological brains are a network of neurons, can help overcome this linear limitation, as will be shown in following paragraph. The first type of network architecture we consider is the feedforward network, in which the  network does not contain cycles. Figure \ref{fig:feedforward} shows a graphical representation.\\
\newpage
In a landmark paper, Cybenko \cite{cybenko1989approximation} proved in 1989 that a feedforward network containing a single hidden layer or neurons with sigmoidal activation could approximate any continuous function of $\mathbb{R} \rightarrow \mathbb{R}$, renewing interest in the field. In 1991 Hornik \cite{Hornik:1991} proved the same for any activation function that is continuous, non-constant, bounded, and monotonically-increasing, showing that it is the multilayer feedforward architecture that provides the universal approximation power of neural networks.

Armed with this theoretical knowledge, research focused on the practical application of feedforward networks. To approximate functions whose analytical form is unknown or non-existent, the proper weights have to be applied to the neuron inputs. The search for these weights is the called the training of the network, and algorithms that perform this task are often called 'learning' or 'teaching' algorithms if applied in a supervised context. The most successful of these learning algorithms is the backpropagation method, discovered by Paul Werbos in 1974 \cite{werbos_beyond_1974} but only popularized by Rumelhart et al. in 1986 \cite{rumelhart1986learning}.\\
The backpropagation algorithm is a variant of gradient-descent.

It requires the activation function used by the neurons of the network to be differentiable. It also requires a cost function that can be written as an average over cost functions for individual training examples.\\
The steps of the algorithm are as follows:
\begin{enumerate}
\item Forward propagation of the input values to obtain the activation of each neuron.
\item Backwards propagation of the error for each node, in order to calculate the delta (weight update) for each weight. Computing the delta is done by using the calculus chain rule to calculate the partial derivative of the error with respect to a  weight.
\item Update each weight according to the gradient and learning rate. 

\item Repeat until the error over the data is below a threshold, or the gradient converges, or for a fixed number of iterations .

\end{enumerate}
The algorithm is made possible by using the chain rule from calculus to iteratively compute gradients for each layer, and relies on 4 equations:
\begin{enumerate}
\item An equation for the error in the output layer: $\delta^L$, a vector whose components are given by:
$\delta^L_j = \frac{\partial C}{\partial a^L_j}\sigma^\prime (z^L_j)$.  C is the cost/error, the $\partial C / \partial a^L_j$ measures how fast the cost is changing as a function of the $j^{th}$ output activation,$z^L$ is the vector of weighted input to the neurons in layer L, the final term $\sigma^\prime (z^L_j)$ measures how fast the activation function is changing at this neuron.
\item An equation for the error $\delta^l$  in terms of the error in the next layer $\delta^{l+1}$ given by:  $\delta^l = ((w^{l+1})^T \delta^{l+1}) \odot \sigma^\prime (z^l)$. Here $\odot$ is the Hadamard product for matrices in which elements are pair-wise multiplied \cite{million2007hadamard}, $(w^{l+1})^T$ is the transpose of the weight matrix for the $l+1^{th}$ layer.
\item An equation for the rate of change of the cost with respect to any bias in the network: $\frac{\partial C}{\partial b^l_j} = \delta^l_j$, where $b^l_j$ is the bias of the $j^{th}$ neuron of layer $l$.
\item An equation for the rate of change of the cost with respect to any weight in the network: $\frac{\partial C}{\partial w^l_{jk}} = a^{l-1}_k \delta^l_j$. This lets us compute the partial derivative in term of quantities $\delta^l_j$ and $a^{l-1}_k$ we already know how to compute.
\end{enumerate}
See \cite{backpropNielsen2014} for an in-depth analysis and proof of the backpropagation algorithm.\\
Since no assumptions are made over the training dataset, it is possible to use batch, stochastic or on-line training. In on-line and stochastic learning, each propagation is followed immediately by a weight update. In batch learning, many propagations occur before updating the weights. On-line learning is used for dynamic environments that provide a continuous stream of new patterns. Stochastic goes through the data set in a random order in order to reduce its chances of getting stuck in local minimum. Stochastic learning is also much faster than batch learning since weights are updated immediately after each propagation. Yet batch learning will yield a much more stable descent to a local minimum since each update is performed based on all patterns.\\
The main limitation of the backpropagation algorithm is the long time needed for proper training. One reason for this can be the vanishing gradient problem.

This fundamental limitation of multi-layered neural networks was first described by Hochreiter in his 1991 thesis \cite{hochreiter1991untersuchungen} and further investigated in two following papers \cite{hochreiter1998vanishing, hochreiter2001gradient}. By observing the gradient of the $l^{th}$ layer of an $L$ layer network, which is given by (in vector form):\\ $\delta^l = \sum^\prime(z^l)(w^{l+1})^T \sum^\prime(z^{l+1})(w^{l+2})^T \cdot \sum^\prime(z^L)\bigtriangledown_a C$. \\
Here, $\sum^\prime(z^l)$ is a diagonal matrix whose entries are the $\sigma^\prime(z)$ values for the weighted inputs to the $l^{th}$ layer. The $w^l$ are the weight matrices for the different layers. And $\bigtriangledown_a C$ is the vector of partial derivatives of C with respect to the output activations. The important factors here are $\sigma^\prime(z)$ which get repeatedly multiplied by the weights $w$. If the absolute value of these products are all smaller than 1, then the absolute value of the gradient will be close to 0. Unfortunately, because the activation functions are generally squashing functions with values converging to $-1$ or $-1$ as they approach $-inf$ and $+inf$ respectively, so as $w$ gets very large or very small, $\sigma^\prime(z) = \sigma^\prime  (wa+b)$ goes to 0. Because each layer adds an extra product to the computation of the gradient, the weights furthest from the output are the slowest to change. This has been verified empirically \cite{backpropNielsen2014, glorot2010understanding, lecun2012efficient}.\\
Despite this issue, deep neural networks have enjoyed success in recent years. Effective pre-training techniques have shown that careful choices in initial weights can lead to efficient learning \cite{sutskever2013importance}, and specialized architectures such as convolutional nets which are biologically inspired networks in which the weights are tied and connections sparse \cite{collobert_unified_2008, krizhevsky2012imagenet}. Nonetheless these techniques remain specialized workarounds for specific tasks and domain, and are not appropriate for all tasks.

\section{Recurrent Neural Networks}
Part of the appeal of neural networks is the parallel with the human brain. However, the network architecture of neurons within the brain is decidedly not a feedforward architecture. In the human brain, billions of neurons are combined in separated lobes and cortices, with electrical signals traveling in all directions. This type of network in which contains cycles are called Recurrent Neural Networks (RNNs). These RNNs are useful because they have superior theoretical computational power. A feedforward network approximates a mathematical function, whereas RNNs approximate dynamical systems. Dynamical systems are essentially functions with an added time component, the same input input can result in a different output at different timesteps. In our case, we consider discrete-time systems, defined by a function $f: X \rightarrow X$, with $X$ the configuration space, and the state of the system evolving from state $x \in X$ to $f(x+1, f(x))$ etc. 

It is the presence of cycles in the network that allow these dynamical changes to occur. In theory, RNNs are capable of "remembering" input values for some time by preserving them in some form within the activations of nodes in the network. A properly trained RNN is capable of learning context sensitive information without having to engineer task-specific data representations as input to the network (e.g. triphones or n-grams in common NLP tasks), though these can still be used with RNNs. Additionally, the study of RNNs is essential to any type of research done on modeling or simulating a human or animal brain. Figure \ref{fig:reservoir} gives a graphical example of a small RNN.
\begin{figure}[!htbp]
\centering
\includegraphics[width=0.3\textwidth]{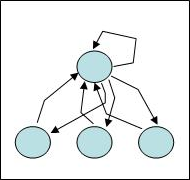}
\caption{Model of a a small Recurrent Neural Network. Image from \cite{rnn} }
\label{fig:rnn}
\end{figure}\\
Of course, the increased complexity of the network architecture comes at a cost. New techniques for training have to be devised (see section on Backpropagation Through Time below),
 and in practical cases are almost always slower than feedforward network training algorithms. The universal approximator theorem must be proven again for RNNs. This important result came in 1991, when Siegelmann and Sontag \cite{Siegelmann91turingcomputability} proved that RNNs, of finite size and with sigmoidal activation function for the nodes, have the capacity to simulate a universal Turing machine, confirming the superior computational power of RNNs over simple feedforward networks.

\subsubsection{Backpropagation through Time}
\begin{figure}[!htbp]
\centering
\includegraphics[width=0.9\textwidth]{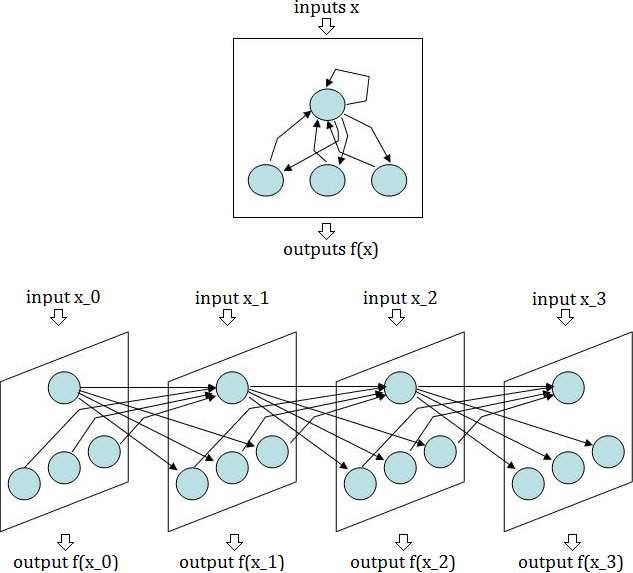}
\caption{Visualization of Backpropagation Through Time unfolding the RNN.}
\label{fig:bptt}
\end{figure}

For practical applications, a successful training technique for RNNs has been Backpropagation Through Time (BPTT), which was independently derived by numerous researchers but popularized in 1990 by Paul Werbos \cite{werbos_backpropagation_1990}. It is an adaptation of the well-known backpropagation training method known from feedforward networks. The feedforward backpropagation algorithm cannot be directly transferred to RNNs because the error backpropagation pass presupposes that the connections between units induce a cycle-free ordering. The solution of the BPTT approach is to "unfold'' the recurrent network in time, by stacking identical copies of the RNN, and redirecting connections within the network to obtain connections between subsequent copies. This gives a feedforward network, which is compatible with the backpropagation algorithm. Figure \ref{fig:bptt} gives a visual explanation of the basic unfolding idea of BPTT.

In this unfolded, feed-forward network, the weights are identical between the copies. The forward pass of one training batch consists of stacking a new layer at each timestep, i.e. at timestep $n$, the current output $f(x_n)$ is computed from the current input $x_n$ and the previous internal state at time $n-1$.
The classical backpropagation algorithm is straightforwardly adapted to this unfolded architecture. The teaching data now consists a single input-output time series. The changes over the tied weights is averaged after each training batch.
The remarks concerning the slow convergence of standard backpropagation carry over to BPTT, even more so since the size the network grows for every iteration. This has limited the used of BPTT to small networks, in the order of tens or hundreds of nodes. Another drawback of this batch training method is that the entire input-output time series must be used, which excludes any kind of online adaptation. A possible solution to both the long computation time and the need for the entire time series is to truncate the past history and use a only $p$  past inputs, in effect doing mini-batches of training. However this limits the memory capacity of network and prevents time dependencies longer than $p$ to be modeled.\\
The shortcomings of the BPTT algorithm hold for other classical RNN training methods.
\vspace{-1.7mm} 
\begin{itemize}[noitemsep]
\item The algorithms are computationally expensive, limiting their use to small networks.
\item Gradient-descent methods suffer from vanishing gradients, making it impossible to guarantee converge. This also leads to dependencies requiring long-term memory becoming hard to learn since gradient information exponentially dissolves over time.
\end{itemize}

\chapter{The Reservoir Computing Paradigm}
In 2001 a fundamentally new approach to RNN design and training was proposed independently by Wolfgang Maass under the name of Liquid State Machines and by Herbert Jaeger under the name of Echo State Networks. This approach is now often referred to as the Reservoir Computing Paradigm. Reservoir computing also has predecessors in computational neuroscience (see Peter Dominey's work in section \ref{dominey}) and in machine learning as the Backpropagation-Decorrelation learning rule proposed by Schiller and Steil in 2005 \cite{steil2004backpropagation}.\\
As Schiller and Steil noticed, when applying BPTT training to RNNs the dominant changes appear in the weights of the output layer, while the weights of the deeper layer converged slowly. It is this observation that  motivates the fundamental idea of Reservoir Computing: if only the changes in the output layer weights are significant, then {\itshape the treatment of the weights of the inner network can be completely separated from the treatment the output layer weights.}

\section{Reservoir Models}
\begin{figure}[!htbp]
\centering
\includegraphics[width=1.0\textwidth]{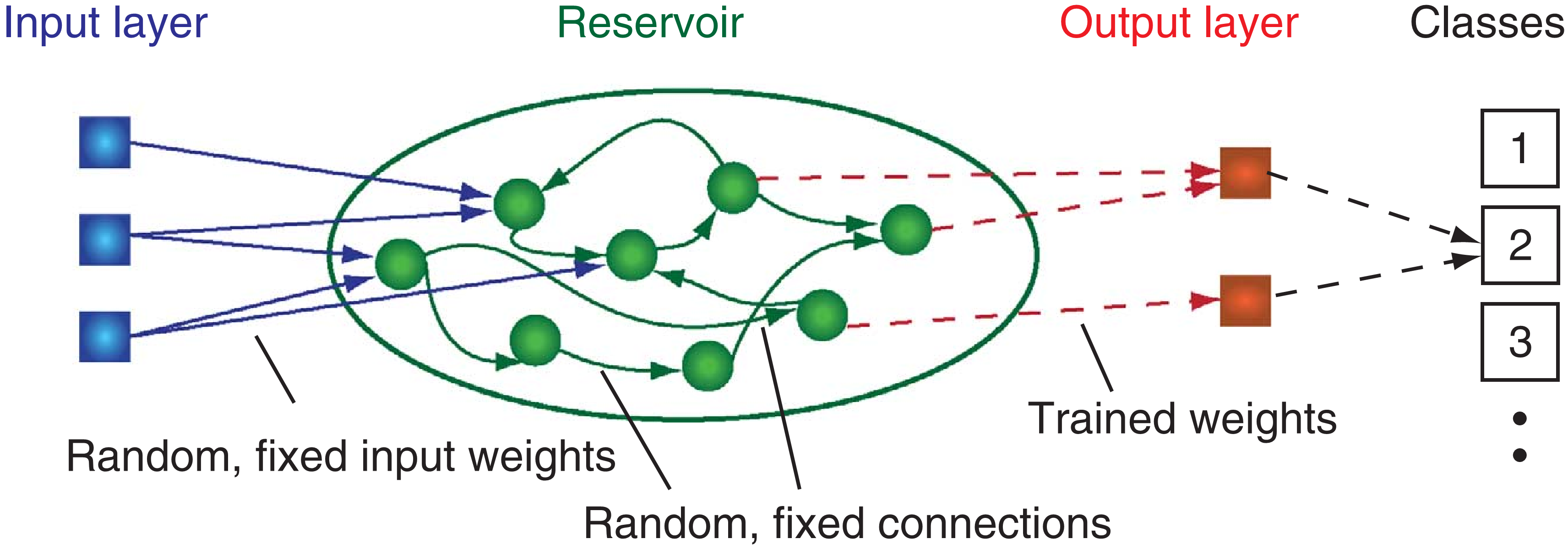}
\caption{Model of a Reservoir Neural Network. Image from \textit{Information Processing using a Single Dynamical Node as Complex System} by Appeltant et al. \cite{appeltant2011information} }
\label{fig:reservoir}
\end{figure}

Reservoir computing methods differ from traditional RNN learning techniques by making a conceptual  and computational separation between the reservoir, i.e. the inner neurons and weights of the network, and the readout, the neurons and weights that produce the output. More specifically, in traditional supervised learning the error between the desired output and computed output will potentially influence the weights of all the network. By contrast, in the Reservoir Computing paradigm, the error will only influence the weights of the readout layer. The weights of the reservoirs connections are set at the start of the learning and do not change. Figure \ref{fig:reservoir} give an graphical overview of the reservoir model.
\\
Despite being a relatively new concept, the reservoir computing paradigm has already been successfully applied to a range of scientific fields, including but not limited to neurosciences, speech processing and robotics. See chapter \ref{lit} for a review of papers using reservoir computing. The following sections detail different brands of reservoir methods: Echo state Network, Liquid State Machines, and other networks that fall into the reservoir category. Most of these techniques were developed independently and have only been united under the term reservoir since 2008. Each brand is a specialized version of reservoir computing, with its own name, history and motivations.

\subsubsection{Echo State Networks}
The Echo State Networks (ESNs) method, created by Herbert Jaeger and his team \cite{jaeger_echo_2001}, represents one of the two pioneering reservoir computing methods. Having observed that if a RNN possesses certain behavioral properties (separability and echo state, see Section \ref{theory}),
 then it is possible to achieve high classification performance on practical applications simply by learning a linear classifier on the readout nodes, for example using logistic regression. The untrained nodes of the RNN are part of what is called the dynamical reservoir, which is where the name Reservoir Computing comes from. The Echo State names comes from the input values echoing throughout the states of the reservoir due to its recurrent nature. Because ESNs are motivated by machine learning theory, they usually use sigmoid neurons over more complicated biologically inspired models.

\subsubsection{Liquid State Machines}
Liquid State Machines (LSMs) are the other pioneer method of reservoir computing, developed simultaneously and independently from Echo State Networks, by Wolfgang Maass \cite{Natschlager2002}. Coming from a computational neuroscience background, LSMs use more biologically realistic models of spiking integrate-and-fire neurons and dynamic synaptic connection models,
 in order to understand the computing power of real neural circuits. As such LSMs also use biologically inspired topologies for neuron connections in the reservoir, in contrast to the randomized connections of ESNs. However, creating a network that mimics the architecture of the cerebral cortex requires some engineering of the weight matrix; and simulating spiking neurons, which are dynamical systems themselves in which electric potential accumulates until it fires leaving a trail of activity bursts, is slow and computationally intensive. Both these facts make the LSM method of reservoir computing more complicated in their implementation  and have not commonly been used for engineering purposes.  

\subsubsection{Other Types of Reservoir Networks}
As mentioned previously, the idea of treating the reservoir and the readout layer separately was also devised by Schiller and Steil \cite{steil2004backpropagation}. They proposed an algorithm called Backpropagation-Decorrelation as a new RNN training method, boasting fast convergence and good practical results, providing a conceptual bridge between traditional BPTT which applies weight update to the whole network, and the reservoir computing paradigm which focus solely on the output layer weights.

We should also mention that Peter Dominey's was probably the first to properly state the reservoir computing principles by observing that 'there is no learning (adaptation) deep within the neural network, and that these connections are randomized (in the pre-frontal cortex in his case)' \cite{DomineyRamus00}. Dominey called these networks Temporal Recurrent Networks, and only recently have computational researchers become aware of the similarities with reservoir networks.\\
The Reservoir Computing paradigm can be extended beyond neural networks. Taking the idea of a reservoir and echoes quite literally, an experiment was set up where the inputs were projected into a bucket of water, and by recording the waves bouncing around the liquid's surface, the authors were able to successfully train a pattern recognizer. Another exotic idea for an untrained reservoir is an E.Coli. bacteria colony, with chemical stimuli as input and protein measures as output. These experiments show that the reservoir computing paradigm may be suitable to harness computational power with unexpected hardware material.

Despite having different names, all the approaches above (ESNs, LSMs, BPDC, Temporal Recurrent) are now referred to as reservoir computing, and we will be using this term throughout the rest of the paper.

\section{Reservoir Computing Theory}
\label{theory} 
While reservoir computing techniques can rely on randomized networks to obtain good performance, there is no guarantee that it is optimal. In fact 'random' is almost by definition the opposite of 'optimal'. While no single technique will work for all tasks, some general rules exist to create reservoirs with good behavior.

\subsubsection{Echo State Property}
An important element for Reservoir Computing to work, coming from the ESNs approach, is that the reservoir should have the echo state property. This condition in essence states that the effect of a previous state and a previous input on a future state should vanish gradually as time passes, and not persist or worse, get amplified. For practical purposes, the echo state property is assured if the matrix W of reservoir weights is scaled so that its spectral radius $\rho(W)$ (i.e., the largest absolute eigenvalue) is close to or inferior to 1. Intuitively, the spectral radius is a crude measure of the amount of memory the reservoir can hold, the small values meaning a short memory, and the large values a longer memory, up to the point of over-amplification when the echo state property no longer holds. Since one incentive to use RNNs is their capacity to have a memory of the inputs, it is important that any method that guarantees the echo state property does not reduce the memory capacity to nothing. A practical test to measure this memory capacity is: can the reservoir output recreate the input, with a k steps delay? 

\subsubsection{Separability Property}
Another important aspect of good reservoirs is the capacity to separate two different inputs, i.e. two different input signals should not have the same output signals. A good heuristic measure of a reservoir's separability power is to compute the distance between different states caused by different input sequences. One method of achieving good separability is to make the matrix defining the network connections sparse, making the activation signals within the network decoupled from each other. Larger networks also lead to more varied activations, and higher separability power.

\subsubsection{Topology}
\label{topo}
The question as to what is the best way to arrange the neurons in the network is still an unresolved one. A larger number of neurons in the network will mean more activation signals, allowing for finer grain classification and higher performance (see \cite{triefenbach_phoneme_2010} for an example on how size can improves performance). But, concerning the particular topology of the network, it has been noted that there is substantial variation in ESN performance among randomly created reservoirs \cite{jiang2008supervised}. After all randomness is almost by definition the opposite of optimal. As mentioned previously, a sparse matrix helps with the performance, but the optimal topology for a reservoir network remains unknown. In a 2004 study \cite{Liebald2004}, different possible topologies were tested, such as scale-free, small world or biologically inspired. Scale-free and small world networks are both widely studied graph topologies that are frequently observed in nature. However, none tested to "perform significantly better than simple random networks'' on two dynamical system prediction tasks. Still the difference in performance amongst these random networks indicates that similar approaches might be useful.

\chapter{How the Reservoir Computing Paradigm is used}
\label{lit}
The universality of the reservoir computing paradigm has lead to its application in a diversity of scientific fields. Since 2009, over a thousand papers have been published on the subject \cite{GS2014}. Among these fields are computational neuroscience \cite{buteneers2009real, bernacchia_reservoir_2011, hinaut_2012, hinaut_exploring_2014}, reservoir computing theory \cite{lukovsevivcius2012reservoir, hermans2012recurrent, chatzis2011echo, oubbati_multiobjective_2012}, speech processing \cite{triefenbach_phoneme_2010,JaDmMjp2013,JaTfDkMjp14} and physical implementation \cite{appeltant2011information, paquot2012optoelectronic, appeltant2014constructing}.
This survey focuses on a select few of these papers with the goals of showing the diversity of reservoir computing applications and discussing the strength and weaknesses of these approaches. 
Lastly, a quick review of a similar randomized network technique, named Extreme Learning Machines \cite{huang_extreme_2006,huang_extreme_2011}, will be given to show the broad scope of the random network principle of reservoir computing.

\section{Neuroscience}
Usage of the Reservoir Computing paradigm in the sciences at large can be classified in two broad categories: use as a powerful machine learning tool, or use as a biologically feasible mechanism. 

\label{dominey}
In research that predates the formulation of the reservoir computing paradigm, Dominey  et al. \cite{Dominey95, Domineyetal03} argue that the brain exhibits reservoir-like processes, i.e. random connection between neurons and linear or simple learning on only the output layer. The learning algorithm described by Dominey is a version of the Least Mean Squares algorithm, in which the output weights are updated by gradient-descent in order to minimize the mean square of the error.

Continuing in this vein of research, in 2011 Nature Neuroscience published a paper by Bernacchia et al \cite{bernacchia_reservoir_2011} which shows how the brain could predict expected rewards over multiple timescales by processing the available information through a reservoir network. By measuring the activity of cortical neurons in monkeys performing a gaming task of matching pennies, the authors observed that some neurons firing indicated that the monkey expected a reward immediately, while others indicated an expected reward in the more distant future. The timescale of these neuronal responses ranged from hundreds of milliseconds to tens of seconds. To replicate this phenomenon, the authors implemented a reservoir neural network of 1000 neuron,a similar size to that measured on the animals. The activity of neurons evolves according to $\frac{dv}{dt} = J \cdot v(t) + h \cdot Rew(t)$ where $v$ is the vector of neuron activity, $J$is the synaptic connectivity matrix of their interactions and $h$is a vector representing the relative strength of the reward input $Rew(t)$ to each neuron. The matrix J of the connection weights was created to be sparse and random. By broadly distributing the connection weights, the authors allows the network to respond to inputs on a wide variety of timescales. In addition, the connection matrix was made to be normal ($J^*J = JJ^*$). This ensures \cite{spectra} that the network dynamics are robust to small changes of the connection strengths. It was then observed that the activity of neurons in the reservoir displayed similar patterns that observed on live monkeys. The paper concluded that animals may be able to process information on multiple timescales thanks to the reservoir's recurrent network capability to maintain multiple memory traces at once.

Partnering with Dominey, the recent work of Hinaut et al. \cite{hinaut_2012, hinaut_exploring_2014} also uses reservoir computing as a explanation to a biological process. Hinaut explores how language is learned, working at the frontiers of neuroscience, natural language processing and robotics. In the paper from May of 2014 \cite{hinaut_exploring_2014} Hinaut and his colleagues argue in that human language can be learned through general associative mechanisms in a stimulus rich environment, in contrast to the Chomsky's innatism that believes that the child's stimulus environment is too poor and that language can only be learned via a highly specialized universal grammar system.  The authors aim to show child's social environment contains enough non-linguistic information that help with the acquisition of phoneme, word, sentence meaning. Through simple interactions with an iCub robot, they showed that the robot is not only capable of learning the grammatical structure of sentences, but also able to produce sentences describing the human's actions. For those two tasks, comprehension and production, the neural language model they used was a random reservoir network with bio-inspired neurons. The reservoir is modeled after the human prefrontal cortex: the reservoir corresponds to the cortex, and the readout layer corresponds to the striatum. The reservoir is composed of leaky neurons with sigmoid activation. The following equation describes the internal update of activity in the reservoir: $x(t + 1) = (1-\alpha)x(t) + \alpha f (W_{res}x(t ) + W_{in}u(t + 1))$ where $x(t)$ represents the reservoir state at time $t$; $u(t)$ denotes the input that time;$\alpha$ is the leak rate; and $f$ is the hyperbolic tangent (tanh) activation function. $W_{in}$ is the connection weight matrix from inputs to the reservoir and $W_{res}$ represents the recurrent connections between internal units of the reservoir. The readout activity is defined by the weight matrix $W_{out}$ which is multiplied to $x(t)$. This readout matrix was trained by linear regression with bias and pseudoinverse method described by Jaeger in 2001 \cite{jaeger_echo_2001}. The results of these experiments is a robot capable of learning that sentences like "John hit Mary'' and "Mary was hit by John'' have the same meaning and same grammatical structure: agent(John), object(Mary), predicate(hit).

\section{Machine Learning}
The reservoir computing paradigm is general and powerful enough to also be applied as tool, rather than as an explanation for natural phenomena. In 2012 Oubbati et al. \cite{oubbati_multiobjective_2012} adapted the reservoir paradigm to multi-objective problems, in which more than one objective function need to be optimized together.
These types of problems are perhaps more naturally occurring than single-objective problems, after all most human decision need to balance multiple goals, for example social/economic, speed/quality. Unfortunately multi-objective problems are also much harder, if not impossible to solve exactly, since the size of all the Pareto (non-dominatied) optimal solutions is potentially infinite. In this paper, the authors apply the reinforcement learning framework to the multi-objective problematic, and use a technique called Adaptive Dynamic Programming (ADP) \cite{wang2009adaptive}, which was developed to solve one of the commons pitfalls of multi-objective reinforcement learning algorithms, the curse of dimensionality. The ADP tool approximates the expected reward using a system called 'Critic'. The reward can be seen as the agent's preference over the objectives, and the expectation is the agent's belief related to how actions impact outcomes. The authors state that the critic is usually a neural network, and requires powerful computational tools to be effective. This is where the reservoir computing paradigm becomes useful. The reservoir estimates several rewards simultaneously and provides their gradients, which are required for the agent to adapt its behavior to the multiple objectives, while being a lot less computationally intensive than traditional neural networks, since reservoir only optimized the output layer of the network.
The authors test their hypothesis by implementing this Reservoir-Computing-ADP in a simulated environments, in which an agent must balance three different objective functions. It was shown that the agent was able to estimates several utilities simultaneously, reaching different Pareto optimal solutions depending on how the objectives were weighted.

\section{Speech Processing}
One of the attractive traits of recurrent neural networks is their capacity to process temporal information. One domain where such temporal information is important is in speech processing. For example, for an automatic speech recognizer to work properly, a sound uttered at time $t$ will influence the recognition probabilities of phonemes both before and after. To allow traditional feed-forward networks to process these temporal dependencies, solutions such as 'context-dependent' phones and N-grams have been engineered. An N-gram is a contiguous sequence of n items extracted from text or speech. These can be phonemes, syllables, letters or words depending on the task. Feeding these N-grams into a feedforward network allows it to process some contextual information. However the range of the context is fixed by preprocessing of the data, and cannot be learned by network. More generally, to avoid the the 'curse of dimensionality', it is assumed that the simplifying Markov property holds, i.e.that future states only depend on the near past.

\begin{figure}[!htbp]
\centering
\includegraphics[width=0.9\textwidth]{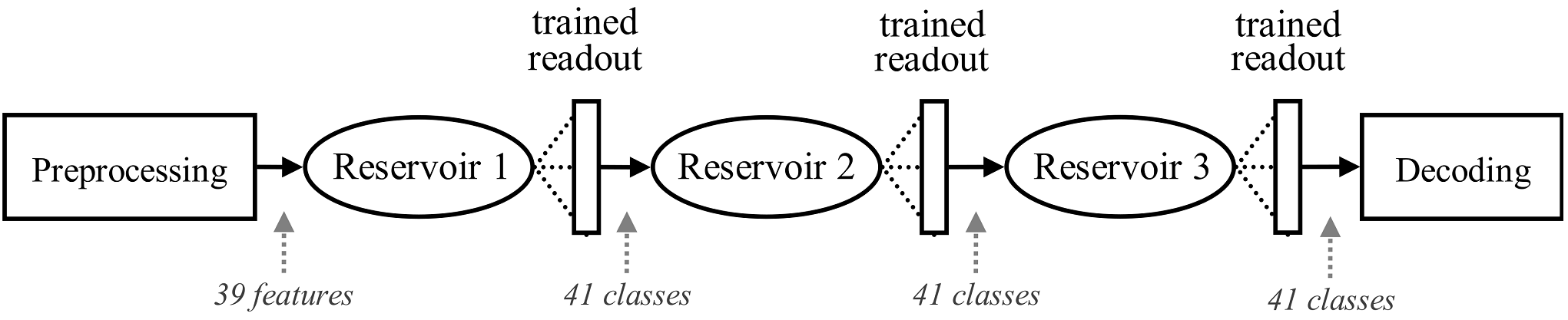}
\caption{The reservoir architecture proposed by Treifenbach et al. Image from \cite{triefenbach_phoneme_2010}}
\label{fig:layered-res}
\end{figure}

The memory capacity of RNNs allows reservoir networks to store this information and use it without needing preprocessing, though it certainly is compatible with such techniques. In the field of natural language processing, reservoir computing research has been led by B. Schrauwen and Jean-Pierre Martens of Ghent University. Amongst the many papers produced by the group (\cite{verstraeten_isolated_2005, jalalvand_connected_2011}), 
an important one for the popularity of the reservoir computing paradigm in the speech recognition domain was published in 2010 by Triefenbach, Jalalvand and the aforementioned Schrauwen and Martens \cite{triefenbach_phoneme_2010}. This particular paper explores the task of continuous phoneme recognition on the popular TIMIT dataset \cite{timit}, with the goal of comparing the performance of single a reservoir network against the novel hierarchical reservoirs architecture, as shown in figure \ref{fig:layered-res}.
\\
The authors argue that additional reservoirs can correct some of the errors made by the previous reservoir. And indeed, in their implementation they observed that the second layer induces a significant improvement of the correction error rate by 3-4\%. They also observed that additional layers beyond the second only provided minimal gain. The overall performance of reservoir networks for the task was comparable to other state of the art systems such as Hidden Markov Models and Deep-Belief nets. 

\section{Physical Implementation}
\begin{figure}[!htbp]
\centering
\includegraphics[width=0.9\textwidth]{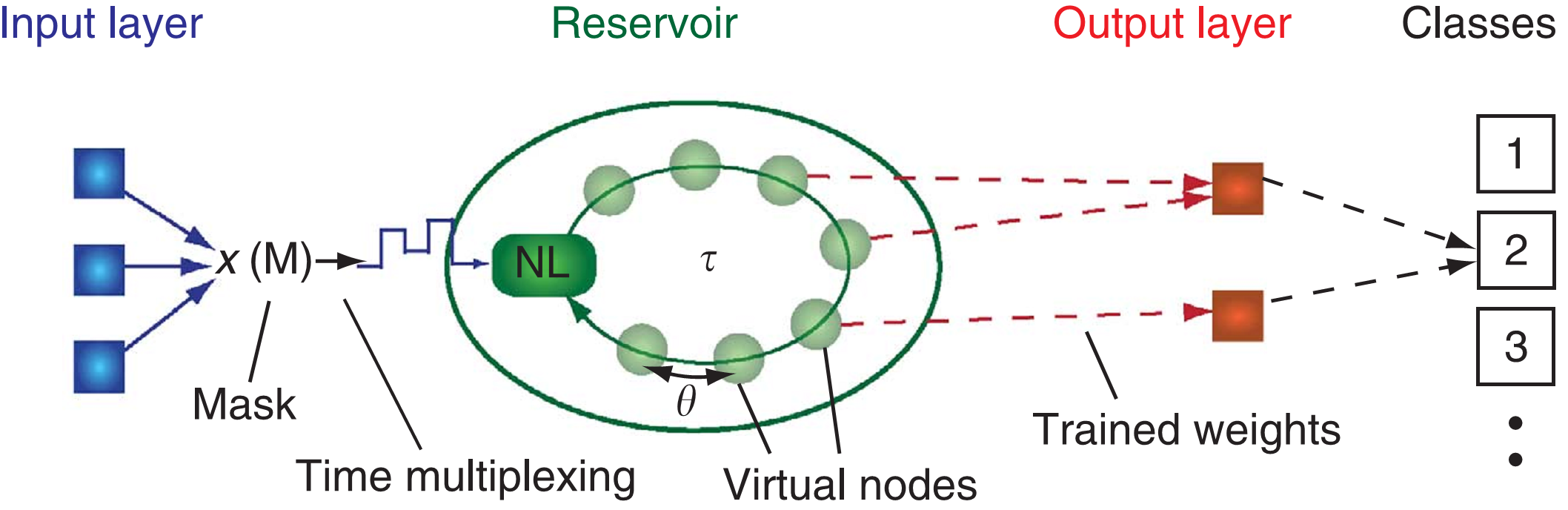}
\caption{Scheme of the Reservoir Computer proposed by Appeltant et al. Image from \cite{appeltant2011information}}
\label{fig:delayed_fb}
\end{figure}

If reservoir networks are to be used in practical applications, implementing them using dedicated hardware would further improve the computational speed of the system over traditional RNNs optimization methods. In recent work published by Nature, Appeltant et al. \cite{appeltant2011information, appeltant2014constructing} 
explore this challenge, with feasibility, performance and resource-efficiency in mind. The authors note that implementing a randomized network architecture, the type of reservoir that is usually coded in simulations, would be a difficult task in hardware. Many components would be needed, one for each reservoir neuron, and the construction process would have to adapt itself every time a new random architecture was devised. In addition to the construction difficulties, not all random networks are efficient, as discussed under \textbf{Topology} in section \ref{topo}, leading to potentially wasted material. To answer these problems, the authors propose to implement a reservoir computer in which the usual reservoir structure of multiple connected nodes is replaced by a dynamical system comprising a nonlinear node subjected to delayed feedback. The delayed feedback creates multiple values that act as 'virtual nodes', whose values are connected to the readout layer, and whose connecting weights are trained using standard reservoir computing procedures. Figure \ref{fig:delayed_fb} gives an overview of the proposed reservoir computer architecture.
\\
For the non-linear node, the authors chose the Mackey-Glass oscillator whose state$X$ evolves according to $\dot X (t) = -X(t) + \frac{\eta \cdot [X(t-\tau) + \gamma \cdot J(t)]}{[1+X(t-\tau) + \gamma \cdot J(t)]^p }$ with $\eta, \gamma, p$ tunable parameters. $J(t)$ is the output of the mask, a preprocessing of the input, where $\tau$ is the delay of the feedback loop, $\theta$ is $\frac{\tau}{N}$ with $N$ the number of virtual nodes. 

The authors tested the validity of this model on two tasks, a spoken digit recognition task and the NARMA task introduced in \cite{atiya2000new} which has become a benchmark for reservoir computing (see \cite{lukosevicius_survey:_2009} for an example). On both the performance is close to or better than that obtained previously with traditional, fully randomized reservoir networks, indicating the potential of this approach.

\section{Other Randomized Networks}
It should be noted that the idea of treating the readout layer separately from the network has also been applied to more traditional feed-forward networks. Under the name of 'Extreme Learning Machines' (ELM), Huang et al. \cite{huang_extreme_2006,huang_extreme_2011} developed a tool very similar to the basic theory of reservoir computing. The weights of the hidden layers of a feedforward network can be randomized, and with proper training of the output weights, good results can be obtained. The authors proves several theorems concerning these ELMs, notably that they are universal function approximators. Toy problems such as noisy function approximation, and real world tasks such as automated medical diagnostic prediction are tackled with success, comparing favorably to traditional feedforward techniques like backpropagation or support vector machines, showing the promise of the Reservoir / ELM approaches to neural network machine learning, with ELM learning non-temporal classifiers and reservoirs approximating dynamical systems with temporal dependencies.

\chapter{Conclusion and Future Work} 
The reservoir computing paradigm has already proven itself as a powerful tool in numerous scientific domains, both at an biological explanation of animal neural processes, and as a computational tool in tasks requiring complex temporal information processing.

\section{Conclusion}
In conclusion:
\begin{itemize}[noitemsep]
\item The simple reservoir computing paradigm states that recurrent neural networks can be efficiently trained by optimizing only the weights connected to the output neurons, leaving the weights of internal connections unchanged. 
\item The reservoir computing paradigm offers a theoretically sound approach to modeling dynamical systems with temporal dependencies. 
\item The paradigm also promises to be much more computationally efficient than traditional RNN approaches that aim at optimizing every weight of the network.
\item Reservoir computing techniques have successfully been applied to classical artificial intelligence problems, providing state-of-the-art performance on engineering problems, and offering explanations on biological brain processes.
\end{itemize}

Despite these advances, reservoir computing is still a new research topic. Because the paradigm is so general, many options are available to research: the type of neuron, the architecture of the network, the training method for the output weights; no consensus exists and is most certainly task-dependent. It also remains to be seen if the paradigm can be applied to large-scale technological ventures, as currently only a small number of research groups are working with reservoirs. Today neuro-scientists focus on which parts of the brain display reservoir-like patterns and how to model these parts; while computational scientists apply the paradigm to new technical fields, exploring and optimizing the different options mentioned above.

\section{Future Work}
As a member of the Speech Lab $@$ Queens College \cite{slqc}, my work centers on computational speech analysis, with a focus on understanding how prosody and intonation communicate information. As an NSF IGERT \cite{igert} fellow, I am also interested in multi-disciplinary work, possibly working with biomedical data. The reservoir computing paradigm in its general classification power is well adapted to multi-disciplinary work.

\begin{figure}[t]
\leftskip-6em
\includegraphics[width=\paperwidth]{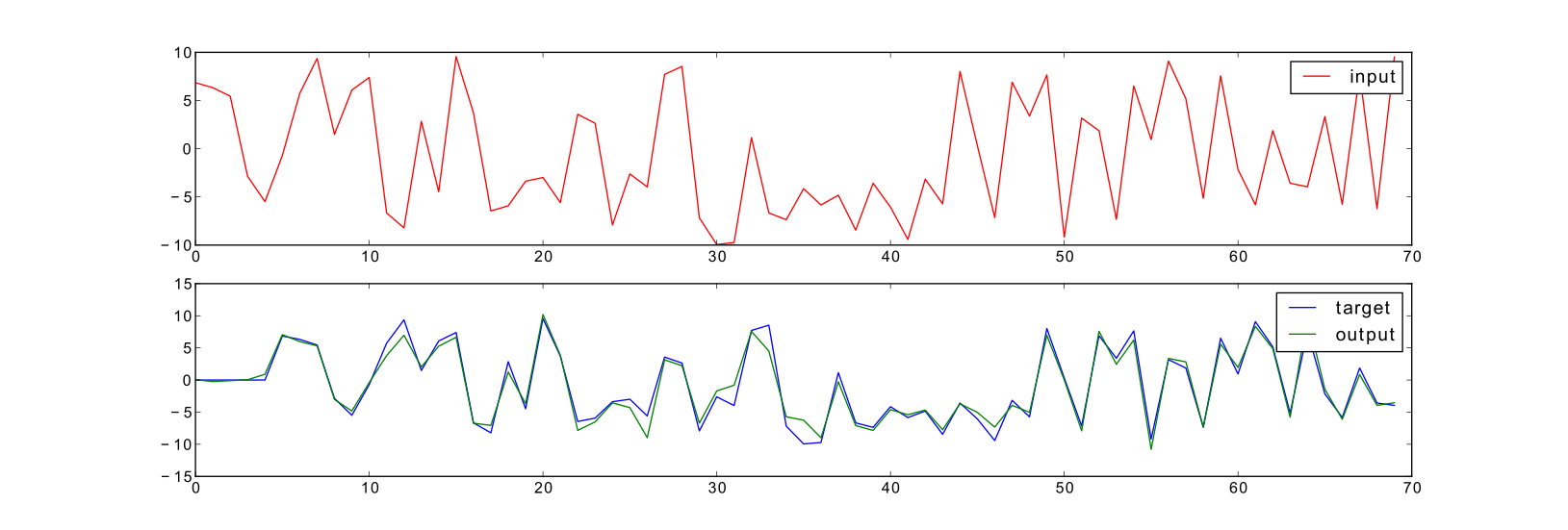}
\caption{Results of a simple 50 neuron random reservoir on a toy task. Note: the results are artificially worsened for visualization
purposes}
\label{fig:myres}
\end{figure}

 I have already explored toy problems using reservoir networks, implemented using the OGER \cite{oger} Python toolbox. Figure \ref{fig:myres} shows my results on a simple task of replicating a noisy input with a time delay of 5 steps, which even a small reservoir is capable of doing perfectly, showcasing the memory capacity of the paradigm.
\\

\backmatter

\bibliographystyle{ieeetr}
\bibliography{bibl}
\addcontentsline{toc}{chapter}{\numberline{}Bibliography }

\end{document}